\documentclass[runningheads]{llncs}
\usepackage{graphicx}
\usepackage{amsmath}
\usepackage{wrapfig,lipsum,booktabs}
\usepackage{amssymb}
\usepackage{caption}
\usepackage[hidelinks]{hyperref}
\usepackage{subcaption}
\usepackage{multirow}
\usepackage{adjustbox}
\usepackage[table]{xcolor}
\def\rvx{{\boldsymbol{x}}}
\newcommand{\myparagraph}[1]{\smallskip\noindent\textbf{#1}}
\begin{document}
\title{Spatial Diffusion for Cell Layout Generation}

\author{Chen Li\thanks{Email: li.chen.8@stonybrook.edu.}\inst{1} \and
Xiaoling Hu\inst{2} \and
Shahira Abousamra\inst{1} \and 
Meilong Xu\inst{1} \and
Chao Chen\inst{1}
}
\authorrunning{C. Li et al.}
\institute{Stony Brook University, Stony Brook, NY, USA \and
Harvard Medical School, Boston, MA, USA\\
% \email{li.chen.8@stonybrook.edu}
}

\maketitle              % typeset the header of the contribution

\begin{abstract}
Generative models, such as GANs and diffusion models, have been used to augment training sets and boost performances in different tasks. We focus on generative models for cell detection instead, i.e., locating and classifying cells in given pathology images. One important information that has been largely overlooked is the spatial patterns of the cells. In this paper, we propose a spatial-pattern-guided generative model for cell layout generation. Specifically, a novel diffusion model guided by spatial features and generates realistic cell layouts has been proposed. We explore different density models as spatial features for the diffusion model. In downstream tasks, we show that the generated cell layouts can be used to guide the generation of high-quality pathology images. Augmenting with these images can significantly boost the performance of SOTA cell detection methods. The code is available at \url{https://github.com/superlc1995/Diffusion-cell}.

\end{abstract}

\section{Introduction}
Cell detection focuses on identifying and locating multiple types of cells in images or videos. Although deep models have achieved great success in cell detection tasks~\cite{abousamra2021multi,hung2020keras,hofener2018deep,yousefi2019transfer,huang2023affine,sugimoto2022multi}, their deployment in the real world is largely constrained due to the demand for large amounts of labeled training data. 
Collecting and annotating cell detection datasets is high-cost.

Despite the rich literature on the generative model, its application in cell detection is still limited. Although GAN or diffusion models can generate high-quality images with a single or a small number of objects, these models cannot generate images with hundreds or thousands of cells. The primary issue is the lack of an explicit modeling of cell spatial distributions. Numerous factors, including cell-cell interactions, morphogenesis, and cellular functions, make the cells follow a specific spatial pattern~\cite{tsai2020adhesion}. 
If a generative model cannot learn these spatial distributions, it will not be able to generate realistic images for data augmentation purposes.
This issue, however, has been largely overlooked by existing generative models. Most existing methods either generate object layouts randomly~\cite{hou2019robust} or use existing layouts directly~\cite{gong2021style}. 

Another constraint for layout generation is the backbone methods. Generative adversarial network (GAN)~\cite{brock2018large,zhang2019self,casanova2021instance,odena2017conditional,miyato2018cgans,abousamra2023topology} cannot handle the large amount of densely packed cells. 
Recent years have witnessed the rise of the diffusion model~\cite{choi2022perception}. Diffusion models outperform GANs in image generation~\cite{dhariwal2021diffusion}, and have shown great potential in many other tasks.
The diffusion model's ability to generate realistic images with fine details makes it ideal for layout generation. 

\emph{We propose a spatial-distribution-guided diffusion model for cell layout generation.}
Our diffusion model learns to generate binary masks, in which each cell is represented by a square marker. 
To incorporate the spatial distribution into the diffusion model, we propose two major ideas. First, to handle the large variation of sparse/dense layouts, we propose to condition the diffusion model on the number of cells. 
% is critical for the generative model to learn and generate reliable object layouts. 
Second, we incorporate spatial distributions into the model. Due to the heterogeneity of spatial distribution, it is unrealistic to simply condition the generation on summary statistics of the spatial density. Instead, we design the model to jointly generate both the layout map and the spatial density maps simultaneously. This way, the model will gradually learn the density distribution through the diffusion process.

As another contribution, we explore and analyze different density distribution models for layout generation: (1) Kernel density estimation (KDE); (2) Gaussian mixture model (GMM); and (3) Gaussian Mixture Copula Model (GMCM)~\cite{tewari2023estimation,kasa2022improved}. While KDE is more flexible, GMM is more constrained and is less likely to overfit. GMCM is a compromise between the two. Our experiments provide a comprehensive analysis of the strengths and weaknesses of the three density models.
Finally, we introduce metric \emph{spatial-FID}, to evaluate the quality of generated object layouts. The metric maps the generated layouts into a spatial representation space and compares them with the spatial representation of real layouts. This provides an appropriate evaluation metric for cell layouts. 

In experiments, we verify the effectiveness of our method through spatial-FID.
Furthermore, we show that the generated layouts can guide diffusion models to generate high quality pathology images. These synthetic images can be utilized as an augmentation to boost the performance of supervised cell detection methods. 

\section{Related work}
\myparagraph{Diffusion Model.}
Diffusion models generate samples from random noise by learning to eliminate the noise in a multi-stage process~\cite{ho2020denoising,sohl2015deep}. Diffusion models are gaining attention because of their superior generation performance compared to GAN models~\cite{dhariwal2021diffusion,brock2018large,karras2019style}.
% \ml{Two sentences start with Diffusion models. Consider replacing the latter one with "They".}
There are extensive applications of diffusion models: semantic segmentation~\cite{graikos2022diffusion,baranchuk2021label}, point cloud generation~\cite{luo2021diffusion}, and video generation~\cite{harvey2022flexible}. Saharia et al.~\cite{saharia2022image} propose the super-resolution method SR3 by conditioning the diffusion model on low-resolution images. Dhariwal et al.~\cite{dhariwal2021diffusion} boost the quality of conditional generation for the diffusion model by using classifier guidance. 
% For object counting, Ranasinghe et al.~\cite{ranasinghe2023diffuse} propose a diffusion based framework for crowd counting, which improves crowd counting performance by predicting density ensemble. 
Zhu et al.~\cite{zhu2022fine} apply the diffusion model to align the crowd distribution for various domains. Unlike previous works, we focus on the cell layout and pathology image generation for cell detection. 
% It is shown that the pathology images with cell layouts generated by our methods have a high quality, and they can be used to boost the performance of existing cell detection methods. 

\myparagraph{Layout Generation.}
Layout generation is the task of synthesizing the arrangement of elements. Generative model based methods~\cite{li2020layoutgan,jyothi2019layoutvae,arroyo2021variational,jiang2022coarse,kikuchi2021constrained,chai2023layoutdm,inoue2023layoutdm} are widely used in this field. LayoutGAN~\cite{li2020layoutgan} and LayoutVAE~\cite{jyothi2019layoutvae} represent pioneering efforts in employing GAN and VAE to generate graphic and scene layouts. LayoutDM~\cite{chai2023layoutdm,inoue2023layoutdm} explores the diffusion based model's capability in layout generation. The works above need access to geometric parameters (location and size) of objects in training layouts. However, only the center coordinates of cells are available for cell detection layouts. TMCCG~\cite{abousamra2023topology} attempts to generate cell layouts with GAN. Considering the superiority of the diffusion model in various generation tasks, we construct a diffusion model based cell layout generation framework. 
% \ml{There's a gap in logic between the last two sentences. TMCCG uses GAN, we use Diffusion, like just to replace the GAN with diffusion. Consider claiming some drawbacks of TMCCG.}

% \myparagraph{Cell Detection.}

\begin{figure*}[t]
    \centering 
   \includegraphics[width=.9\linewidth]{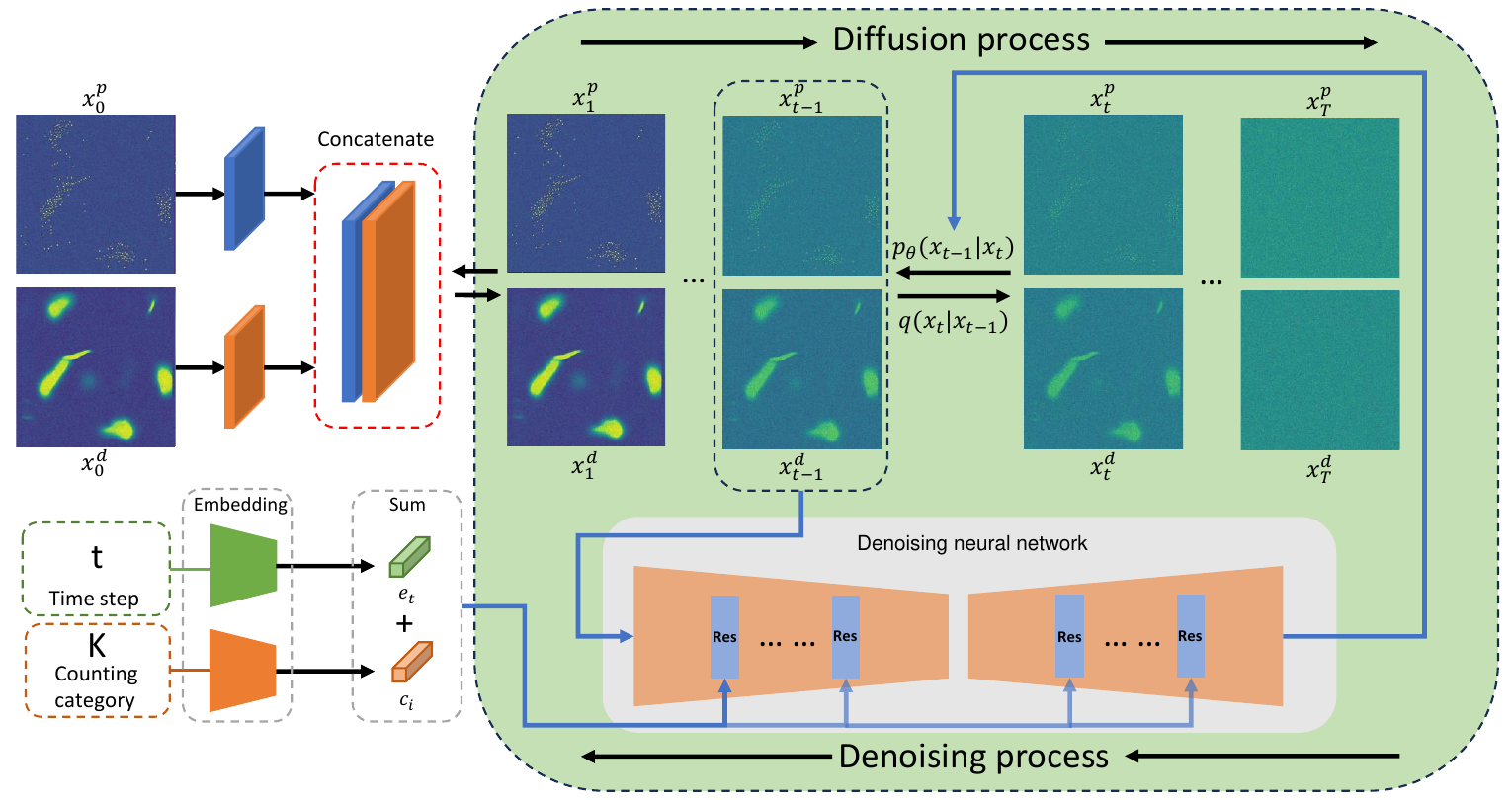}
  % \begin{subfigure}{0.45\textwidth}
  %  \includegraphics[width=1\linewidth]{fig_1_b.pdf}
  % \end{subfigure}
  % \begin{subfigure}{0.45\textwidth}
  %  \includegraphics[width=1\linewidth]{fig_1_c.pdf}
  % \end{subfigure}
\caption{Overview of proposed layout generation framework. In order to reduce the difficulty of learning spatial distributions behind object layouts, we propose a counting-category conditioned object layout generation framework. Meanwhile, the spatial density maps are incorporated into the training and generation process to help diffusion model learn and generate the spatial distribution of object layouts. For clearance, we only show one cell type's cell layout and spatial density map. $\rvx_t^p$ and $\rvx_t^d$ are the cell layout and spatial density map at time step $t$.
% \sa{Please add description of $x^p$ and $x^d$ either in the caption or on the figure. I understand that $x^p$ is a binary map. Why is it not colored in black and white?} 
}
\label{fig:layout_gen}
\end{figure*}

\section{Method}
It is essential to generate realistic pathology images for cell detection tasks, and one challenge is to properly model the spatial distribution of these cell layouts. This paper explores multiple ways to summarize the spatial context in cell layouts. By incorporating spatial information in the training process, diffusion models can learn the underlying distribution of cells and generate realistic layouts. We represent the cells in pathology images as a series of $3\times3$ square markers in layout maps. The markers are in the center of cells. Using narrow markers can prevent layout maps of highly dense cells from collapsing into a big mass. The generated layouts can guide the generation of realistic pathology images.

\subsection{Layout and Image Generation}\label{diff_model}
In this part, we start by explaining our approach to training spatial-aware diffusion models to generate realistic cell layouts. Next, we introduce the diffusion framework for generating synthetic pathology images under the guidance of generated layouts. 

\myparagraph{Diffusion Model.}\label{lay_gen}
% Here we introduce Denoising diffusion probabilistic models (DDPM) briefly.\sh{The introduction of DDPM can be a separate background section and here only focus on how it is used/modified.} 
Given data $\rvx_0  \thicksim q(\rvx_0)$, the forward diffusion process of diffusion models corrupts original data $\rvx_0$ into $\rvx_T$ by introducing Gaussian noise with variance schedule $\beta_1,\dots,\beta_T$. This process is depicted as a Markov chain formulated as follows: 

\begin{gather}
    q(\rvx_1,\dots,\rvx_T|\rvx_{t-1}) := \prod^T_{t=1}q(\rvx_t|\rvx_{t-1}) \\
    q(\rvx_t|\rvx_{t-1}) := \mathcal{N}(\rvx_t; \sqrt{1-\beta_t}\rvx_{t-1}, \beta_t \boldsymbol{I} ) 
\end{gather}
% Besides, given the property of the Markov chain, the sample generated by the diffusion process at the time step $t$ can be sampled by concise form:
% \begin{equation}
%     \rvx_t = \sqrt{\Bar{\alpha}_t}\rvx_0 + \sqrt{1-\Bar{\alpha}_t} \boldsymbol{\epsilon},  \boldsymbol{\epsilon} \thicksim \mathcal{N}(0, \boldsymbol{I})
% \end{equation}
% where $\alpha_t := 1 - \beta_t$ and $\Bar{\alpha}_t := \prod_{s=1}^{t}(1-\beta_s)$. 
If $T$ is large enough, the corrupted data $\rvx_T$ will nearly follow an isotropic normal distribution. 
% In practice, $T$ is set as 1000. 
The diffusion models sample data from $q(\rvx_0)$ by reverse of the diffusion process. The generative process $q(\rvx_{t-1}|\rvx_{t})$ is approximated by a neural network:

\begin{gather}
    p_\theta(\rvx_{t-1} | \rvx_t) = \mathcal{N}(\rvx_{t-1}; \boldsymbol{\mu}_\theta(\rvx_t, t), \sigma^2_t \boldsymbol{I} ) \\
    \boldsymbol{\mu}_\theta(\rvx_t, t)  =  \frac{1}{\sqrt{\alpha_t}} \left (\rvx_t - \frac{1-\alpha_t}{\sqrt{1- \Bar{\alpha}_t}} \boldsymbol{\epsilon}_\theta(\rvx_t, t) \right )
\end{gather}
where $\boldsymbol{\epsilon}_\theta(x_t,t)$ is infered by a denoising neural network. Both $\boldsymbol{\epsilon}_\theta(x_t,t)$ and $\sigma_t$ are learned by optimizing a hybrid learning objective~\cite{choi2022perception,nichol2021improved}. 
% To fulfill the reverse process and generate data from distribution $q(\rvx_0)$, we can use the formula,
% $    \rvx_{t-1} = \boldsymbol{\mu}(\rvx_t,t) + \sigma_t \boldsymbol{\epsilon}$. 
% \sh{up until here can be moved to a background section}

% Due to the sparsity of point cloud maps, it is challenging 
% % \cc{Casual word. Use something formal: challenging} \cl{Changed} 
% for diffusion models to learn the underlining distribution directly from point cloud maps, which prevents diffusion models from generating high-quality layouts 
Due to the sparsity of cell layout maps, it is challenging for diffusion models to learn the underlying distribution of cells from the standard training process. To address this problem, we introduce spatial density maps. By modeling the spatial information in a dense way, the spatial density maps teach the model the spatial distribution behind cell layouts. The bond between layout maps and 
% spatial maps assures the generated cell layouts follow the 
density maps assures the generated cell layouts follow the 
spatial distribution of the training set. To incorporate the spatial density map into the training process efficiently, we construct $\rvx_0$ by concatenating cell layout map $\rvx_0^p$ and spatial density maps $\rvx_0^d$ together: $\rvx_0 = concat(\rvx_0^p, \rvx_0^d)$. We construct $\rvx_0^p$ and $\rvx_0^d$ by combining the cell layouts and spatial density maps of different cell types. We generate spatial density maps for each cell type independently. As shown in Fig.~\ref{fig:layout_gen}, the generated cell layouts are distributed in the corresponding spatial density generation pattern. More samples are available in the supplementary. 

The spatial distribution of cells in pathology images is influenced by density. To make the generation model realize this correlation, we incorporate cell counting in the image patch as prior information for layout generation. However, due to the lack of samples for each quantity, using cell counting as a condition directly will lead to poor-quality layout generation. Here, we split the training image patches by the number of cells in each patch. We define $C$ as the cell count distribution of the training set, and $C_p$ is the $p$-th quantile of this distribution. We can divide the cell counting space into $K \in \mathbb{N}^*$ parts: $[C_{i/K}, C_{(i+1)/K}], i\in \{0\dots K-1\}$. During training, we treat the layouts belonging to the part $i$ as a counting category $e_i$ and condition the diffusion model on these counting categories. In the inference stage, we sample the same number of layouts from each counting category so that the generated layouts share the same counting distribution as the original data. An overview of layout generation and learning process is in Fig.~\ref{fig:layout_gen}.

\begin{figure*}[t]
    \centering 
   \includegraphics[width=1\linewidth]{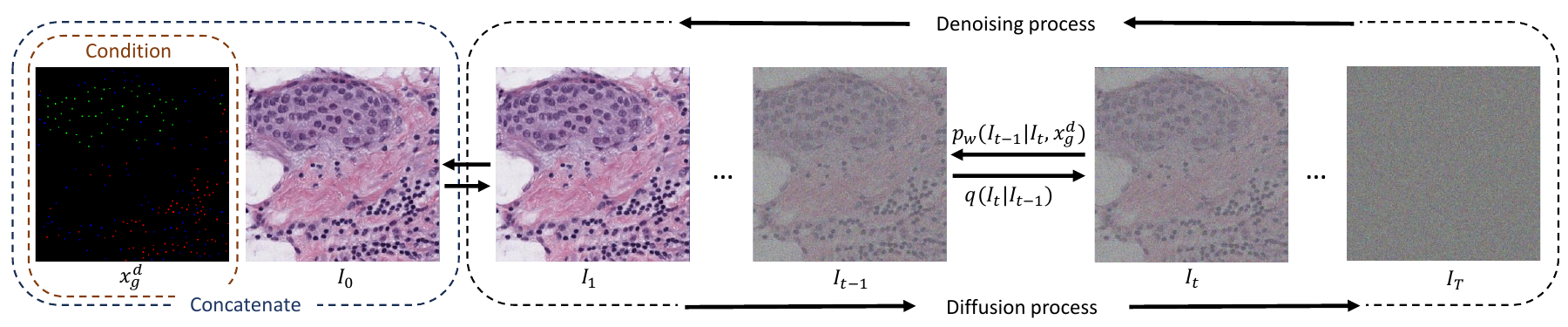}
  % \begin{subfigure}{0.45\textwidth}
  %  \includegraphics[width=1\linewidth]{fig_1_b.pdf}
  % \end{subfigure}
  % \begin{subfigure}{0.45\textwidth}
  %  \includegraphics[width=1\linewidth]{fig_1_c.pdf}
  % \end{subfigure}
\caption{Illustration of cell layout conditioned diffusion model for generating pathology images with ground truth. 
}
\label{fig:img_gen}
\end{figure*}

We generate pathology images $I$ using a diffusion model conditioned on the generated layouts. The layout maps $\rvx_g^p$ are fed into the denoising neural network as a condition to guide the cell distribution of the generated pathology image. Our layout conditional diffusion model $p(I |\rvx_g^p)$ generates pathology images with the layout map $\rvx_g^p$ through the neural network approximated denoising process $p_w(I_{t-1}|I_t, x_g^p)$, where $w$ is the parameter of neural network for pathology image generation. An illustration is shown in Fig.~\ref{fig:img_gen}.
% \sa{Figure number should be ref command and not hard coded. There is no Fig.3 so far.}
Given the generated high-quality pathology images and their cell layouts, we can augment the existing cell detection methods for better performance. 

\subsection{Spatial Feature Extractors}
An excellent spatial feature extractor should be able to capture the cluster patterns in cell layouts, including the position, density, and area of cell clusters, which is a challenging task. Here, we present three different ways to extract spatial features: Kernel density estimation (KDE), Gaussian Mixture Model (GMM)~\cite{reynolds2009gaussian}, and Gaussian Mixture Copula Model (GMCM)~\cite{tewari2023estimation}. 

Kernel density estimation is a non-parametric probability estimation framework, which is widely used for point pattern analysis. Intuitively, KDE treats training data points as density sources, and the combination of effects from training data points will create a smoothed estimate of probability distribution. For each estimating location, the closer data points have higher density contributions. The bandwidth is a crucial parameter for KDE. A lower bandwidth results in an estimation with more details but potentially introduces extra noise. On the other hand, a larger bandwidth produces a smoother result with the risk of oversmoothing. Here, to reach a good balance, we use Scott’s Rule to select bandwidth~\cite{scott:1992}. 

Unlike KDE, the Gaussian mixture is a parametric model, modeling the distributions of object cluster patterns as Gaussian distributions. Therefore, GMM has better interpretability with limited parameters. Here, we introduce another parametric model -- the Gaussian mixture copula model (GMCM). The copula function is used to capture the dependency between marginal densities. GMCM uses GMM to estimate the marginal distribution of point cloud and Gaussian mixture copula function, derived from a mixture of Gaussians, to capture the marginal dependency. Due to the non-identifiability nature of the components in GMCM and GMM, we adopt an EM algorithm~\cite{tewari2023estimation} to optimize the parameters of GMCM and GMM. Bayesian Information Criterion (BIC) decides the optimal number of components for both GMM and GMCM.

\section{Experiment}
% \sa{There are no qualitative results. Can you add some? Show both generated layout map and image for different cell counts.}
We evaluate the effectiveness of cell layout and pathology image generation frameworks. We propose spatial-FID as a Fréchet Inception Distance (FID) modification to measure the quality of generated cell layouts. We also evaluate the effectiveness of our layout conditional pathology image generation framework on cell detection tasks. 
% \sa{This should be cell layout and cell detection.}

\myparagraph{Datasets.} We evaluate our methods on cell detection dataset BRCA-M2C~\cite{abousamra2021multi}. BRCA-M2C has 80, 10, and 30 pathology image patches for training, validating, and testing, respectively. This dataset has three cell types: tumor, lymphocyte, and stromal. 
% It is shown that more realistic generative layouts can lead to higher-quality pathology images and boost the performance of cell detection methods.
% \sa{Add dataset description such patch size, count, cell classes, etc.}

\myparagraph{Implementation details.}
We separate the training set into five counting categories. To ensure the image patches for training have abundant spatial structures to learn, we crop $464 \times 464$ patches from whole training images for training diffusion models on BRCA-M2C. The diffusion model of layout generation is a U-Net architecture~\cite{dhariwal2021diffusion} with multiple attention heads at resolution: $32\times 32$, $16 \times 16$, and $8\times 8$. For the image generation framework, we initialize the model with weights from pre-trained super-resolution diffusion model~\cite{dhariwal2021diffusion}. 
% \sa{What about the architecture and embedding size of the time step and count encoders?} 
For cell detection tasks, we use the state-of-the-art methods MCSpatNet~\cite{abousamra2021multi} and U-Net~\cite{ronneberger2015u} as the cell detection frameworks.

% \begin{wraptable}{r}{6.3cm}
% \caption{A wrapped table going nicely inside the text.}\label{wrap-tab:1}
% \begin{tabular}{c|cccc}
% \hline
% Method & w/o density & GMM   & KDE   & GMCM  \\
% \hline
% Spatial-FID    & 1.078       & 0.039 & 0.357 & 0.953 \\
% % \hline
% % PSU     & 12.390      & 7.827 & 4.264 &       \\
% \hline
% \end{tabular}
% \end{wraptable} 

\myparagraph{Evaluation metric.}
To evaluate the quality of generated layout maps, we propose spatial-FID $(\downarrow)$\footnote{The lower the better.}. The FID is used to measure the performance of the generative model on natural image datasets, e.g., ImageNet and CIFAR-10/100. It thus relies on the visual features from a pre-trained inception model. However, there is a significant difference between natural images and layout maps. Therefore, we train an autoencoder (Fully Convolutional Networks) on the layout maps from the training set to capture the spatial information in the layouts. We use the spatial feature from the middle of the encoder ($f_s(\cdot)$) to replace the visual features. 
Moreover, introduce spatial-FID formed as follows:

\begin{equation}
    s^2(( \mu_T, \Sigma_T), (\mu_D, \Sigma_D)) = 
    \\  \| \mu_T - \mu_D \|_2^2 +Tr(\Sigma_T + \Sigma_D - 2( \Sigma_T\Sigma_D)^{(1/2)})
\end{equation}
where $\mu_T$ ($\mu_D$) and $\Sigma_T$ ($\Sigma_D$) are the mean and covariance of the spatial features extracted by $f_s(\cdot)$ from training layouts (generated layouts), respectively. 
% We call the FID after modification spatial-FID.
We use F-scores to evaluate the cell detection performance.
\setlength{\tabcolsep}{10pt}
\begin{table}[t]
\caption{Quantitative results evaluated by spatial-FID.}\label{fid_res}
\small
\begin{center}
\begin{tabular}{c|cccc}
\hline
Method & w/o density & GMM   & KDE   & GMCM  \\
\hline
Spatial-FID    & 1.078       &\bf 0.039 & 0.357 & 0.953 \\
\hline
\end{tabular}
\end{center}
\end{table}
\setlength{\tabcolsep}{10pt}
\begin{table}[t]
\caption{Results (F-scores) on the augmentation of cell detection. Columns ``Infl.'', ``Epi.", ``Stro.", and ``Det." are F-scores on inflammatory, epithelial, stromal, and all cells.}
% \ml{The caption seems not that accurate. Results of cell detection after extending the training data.}
\label{cell_aug}
\begin{center}
\begin{tabular}{c|c|c|c|c|c}
\hline
Method         & Infl. & Epi.  & Stro. & Mean  & Det 
\\
\hline
U-Net          & 0.498 & 0.744 & 0.476 & 0.572 & 0.838    \\
U-Net + Rand.  & 0.625 & 0.735 & 0.472 & 0.611 & -     \\
U-Net + TMCCG  & 0.650 & 0.768 & 0.511 & 0.644 & -     \\
U-Net + No den & 0.641 & 0.784 & 0.537 & 0.654 &\bf 0.856 \\
U-Net + KDE    & 0.642 &\bf 0.800 & 0.545 & 0.662 & 0.853 \\
U-Net + GMM    & 0.647 & 0.797 &\bf 0.554 &\bf 0.666 & 0.853 \\
U-Net + GMCM   &\bf 0.658 & 0.786 & 0.536 & 0.660 & 0.852 \\
\hline
MCSpatNet          & 0.635 & 0.785 & 0.553 & 0.658 & 0.849 \\
MCSpatNet + Rand.  & 0.652 & 0.772 & 0.506 & 0.644 & -     \\
MCSpatNet + TMCCG  &\bf 0.678 & 0.800 & 0.522 & 0.667 & -     \\
MCSpatNet + No den & 0.647 & 0.788 & 0.543 & 0.659 & 0.850 \\
MCSpatNet + KDE    & 0.652 & 0.793 & 0.556 & 0.667 & 0.845 \\
MCSpatNet + GMM    & 0.639 & 0.804 &\bf 0.563 &\bf 0.669 &\bf 0.855 \\
MCSpatNet + GMCM   & 0.615 &\bf 0.806 & 0.555 & 0.659 & 0.850 \\
\hline
\end{tabular}
\end{center}
\end{table} 
% \setlength{\tabcolsep}{6pt}
% \begin{table}[t]
% \caption{Quantitative results evaluated by spatial-FID.}\label{fid_res}
% \small
% \begin{center}
% \begin{tabular}{c|cccc}
% \hline
% Method & w/o density & GMM   & KDE   & GMCM  \\
% \hline
% Spatial-FID    & 1.078       &\bf 0.039 & 0.357 & 0.953 \\
% \hline
% \end{tabular}
% \end{center}
% \end{table}
\subsection{Results}
In this part, we train the model on the training set, generate 200 layouts with corresponding pathology images for each counting category, and evaluate our method from layout generation and cell detection augmentation.

As shown in Tab.~\ref{fid_res}, the spatial density map generated by GMM achieved the best performance due to the distribution of cells in BRCA-M2C conforming to the mixture of Gaussian best. The cell layout distribution of BRCA-M2C data has distinct subgroups or clusters, and GMM can effectively capture these patterns. The flexibility of KDE leads to noisy density estimation results in our case, preventing our framework from getting better layout generation. GMCM models dependencies between variables using copulas, which can introduce additional complexity. The mismatches between GMCM's assumption and data distribution lead to inferior layout generations.
% \ml{Why can we say that the distribution of cells in BRCA conforms to the GMM best? Can this be the reason, not due to the performance being the highest?}
According to Tab.~\ref{cell_aug}, the pathology image generations directed by GMM boost the cell detection performance best, reflecting that better cell layouts lead to higher-quality pathology image generations. As shown in Fig.~\ref{fig:qualitative}, our pathology image generations are deceptively realistic. 
% \ml{As shown in Fig.~\ref{fig:qualitative}, our pathology....}
This is mainly attributed to high-quality cell layout generations and the excellent performance of diffusion models in pathology image generation.

\begin{figure*}[t]
    \centering 
   \includegraphics[width=1\linewidth]{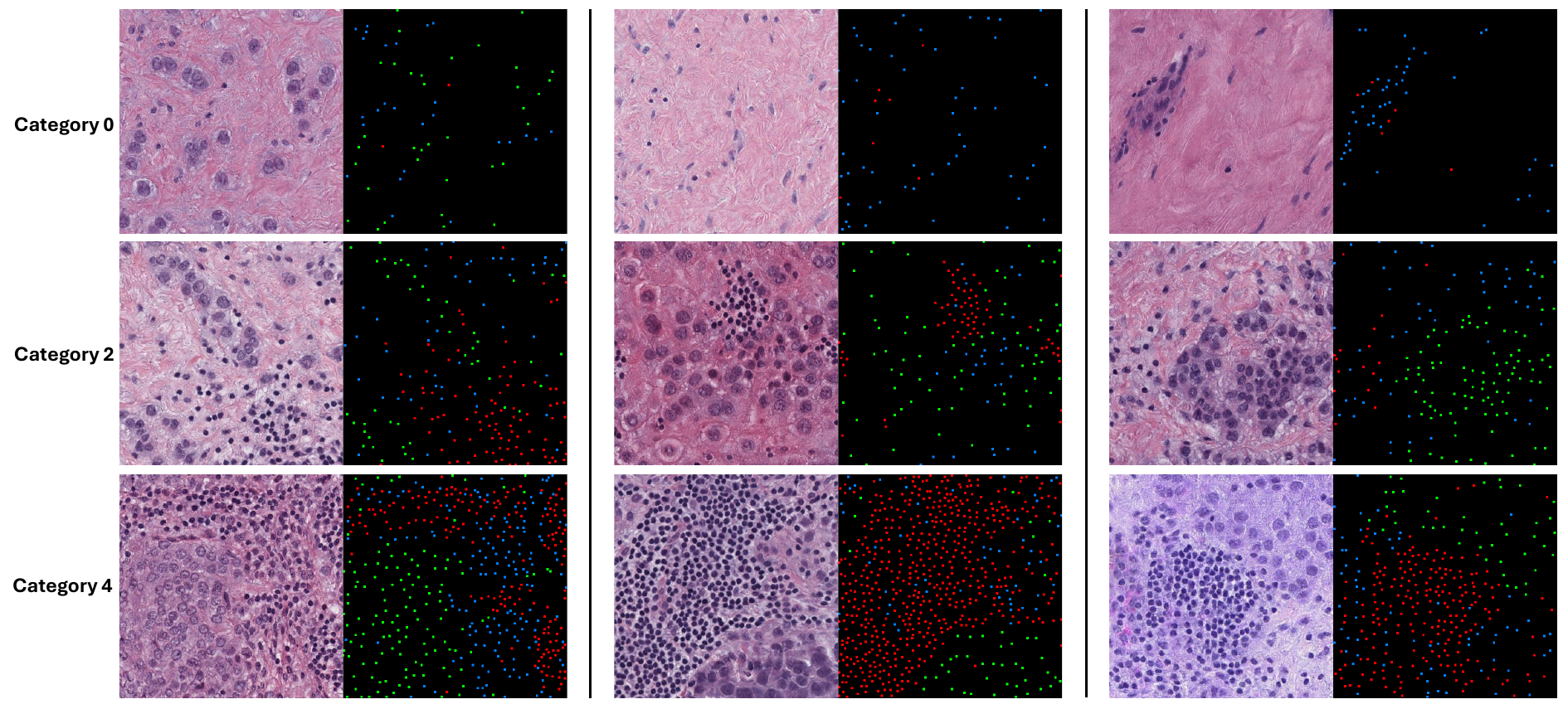}
  % \begin{subfigure}{0.45\textwidth}
  %  \includegraphics[width=1\linewidth]{fig_1_b.pdf}
  % \end{subfigure}
  % \begin{subfigure}{0.45\textwidth}
  %  \includegraphics[width=1\linewidth]{fig_1_c.pdf}
  % \end{subfigure}
\caption{Qualitative results generated by our layout and image generation framework for cell detection.
Rows 1st, 2nd, and 3rd are from the counting categories 0, 2, and 4, respectively. Tumor, lymphocyte, and stromal cells are marked by green, red, and blue marks.
}
\label{fig:qualitative}
\end{figure*}
% \myparagraph{BRCA-M2C}
\subsection{Ablation study}
We conduct ablation studies to show each component's effectiveness and the effects of hyper-parameters on the generation framework. The supplementary includes more ablation studies.

\myparagraph{Counting categories.}
We conduct ablation studies by discarding this condition or setting different numbers of counting categories (10 or 20) to show the effectiveness of counting categories in cell layout generation. According to Tab.~\ref{cc_ab}, counting categories are important for generating realistic cell layouts, and our method is robust to the change of counting categories. 
\setlength{\tabcolsep}{10pt}
\begin{table}[h]
\caption{Ablation study results on the number of counting categories}\label{cc_ab}
\small
\begin{center}
\begin{tabular}{c|cccc}
\hline
Classes number & 0 & 5   & 10   & 20  \\
\hline
Spatial-FID    & 0.122      &\bf 0.039 & 0.231 & 0.558 \\
\hline
\end{tabular}
\end{center}
\end{table}
\section{Conclusion}
In this paper, we propose a spatial-distribution-guided diffusion framework for generating high-quality cell layouts and pathology images. 
To represent the spatial distribution of cell layouts properly, we explore three alternative tools for spatial feature extraction: KDE, GMM, and GMCM. 
They all can significantly boost the generative quality of cell layouts. Due to the underlying cell layouts of BRCA-M2C complying with the mixture of Gaussian best, GMM achieves the best performance. 
We treat generated cell layouts as conditions for pathology image generation. 
These high-quality generated pathology images can improve the performance of SOTA cell detection methods.

\section{Supplementary}

\subsection{Ablation study}
\subsubsection{Bandwidth.}
A good bandwidth selection is critical for the quality of spatial information represented by KDE. Here, we study the effect of bandwidth using constant values. As shown in Tab.\ref{band_ab}, the Scott method is more valid for extracting spatial information from cell layouts.

\setlength{\tabcolsep}{6pt}
\begin{table}[ht]
\small
\begin{center}
\begin{tabular}{c|ccccc}
\hline
Bandwidth & Scott & 0.05 & 0.1 & 0.5 & 1.0   \\
\hline
Spatial-FID    &\bf 0.357  & 1.41 & 0.538 & 0.908 & 1.433 \\
\hline

\end{tabular}
% \vspace{-.1in}
\caption{Ablation study results on the bandwidth of KDE}\label{band_ab}
\end{center}
% \vspace{-.2in}
\end{table}

\subsubsection{Number of components}

As we show in Tab.~\ref{gmm_ab}, selecting the component number of GMM by BIC is essential for generating high-quality spatial density maps by GMM.

\setlength{\tabcolsep}{6pt}
\begin{table}[ht]
\caption{Ablation study results on the component number of GMM}\label{gmm_ab}
\small
\begin{center}
\begin{tabular}{c|ccccc}
\hline
Component number & BIC & 4 & 6 & 8 & 10  \\
\hline
Spatial-FID    &\bf 0.039  & 0.507 & 0.464 & 0.365 & 0.567 \\
\hline
\end{tabular}
\end{center}
\end{table}

\subsection{Samples}

Here, we show more generated layouts, corresponding generated pathology images, and density maps. As shown in Fig.~\ref{fig:qualitative}, the density map can reflect the spatial distribution of cells well.

\begin{figure*}
% \vspace{-.1in}
\centering

  \begin{subfigure}{0.15\linewidth}
   \includegraphics[width=1\linewidth]{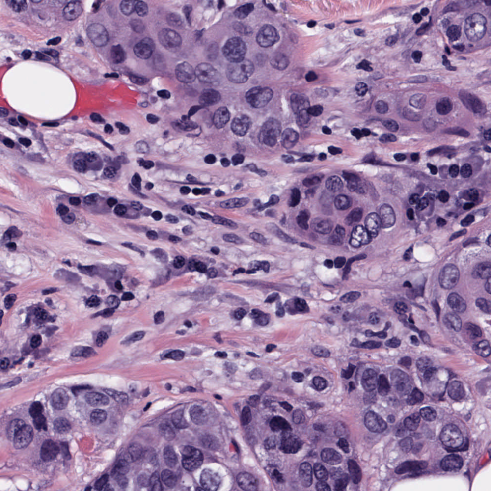}
  \end{subfigure}
  \begin{subfigure}{0.15\linewidth}
   \includegraphics[width=1\linewidth]{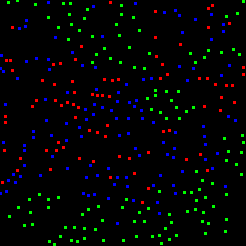}
  \end{subfigure}
  \begin{subfigure}{0.15\linewidth}
   \includegraphics[width=1\linewidth]{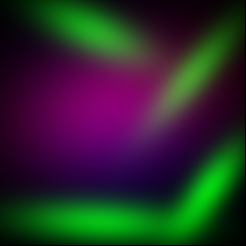}
  \end{subfigure}
      \begin{subfigure}{0.15\linewidth}
   \includegraphics[width=1\linewidth]{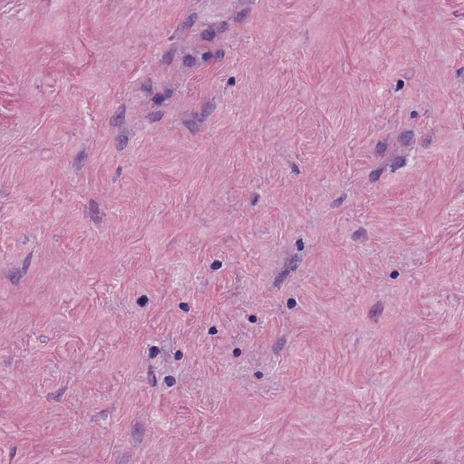}
  \end{subfigure}
    \begin{subfigure}{0.15\linewidth}
   \includegraphics[width=1\linewidth]{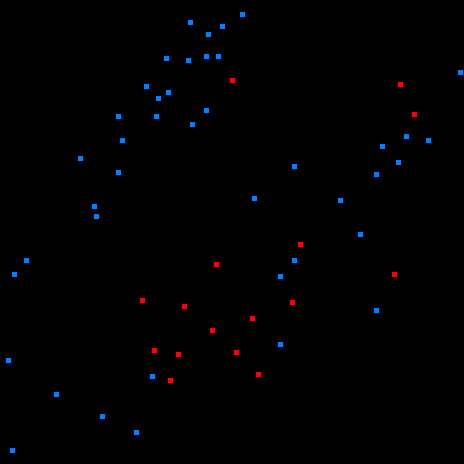}
  \end{subfigure}
    \begin{subfigure}{0.15\linewidth}
   \includegraphics[width=1\linewidth]{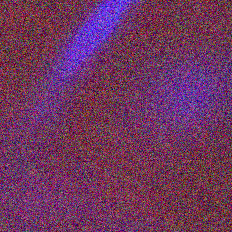}
  \end{subfigure}

  \begin{subfigure}{0.15\linewidth}
   \includegraphics[width=1\linewidth]{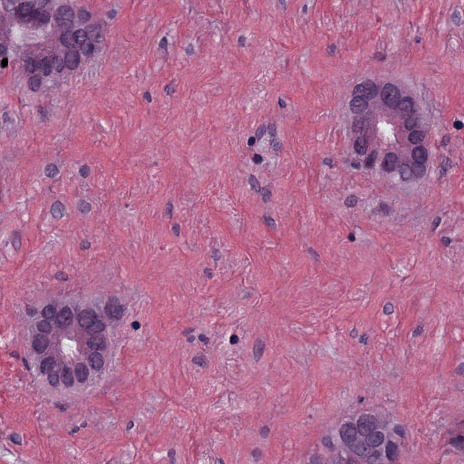}
  \end{subfigure}
  \begin{subfigure}{0.15\linewidth}
   \includegraphics[width=1\linewidth]{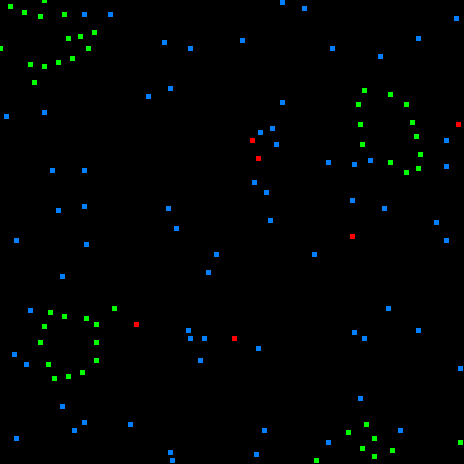}
  \end{subfigure}
  \begin{subfigure}{0.15\linewidth}
   \includegraphics[width=1\linewidth]{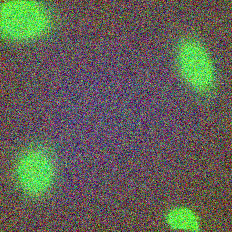}
  \end{subfigure}
      \begin{subfigure}{0.15\linewidth}
   \includegraphics[width=1\linewidth]{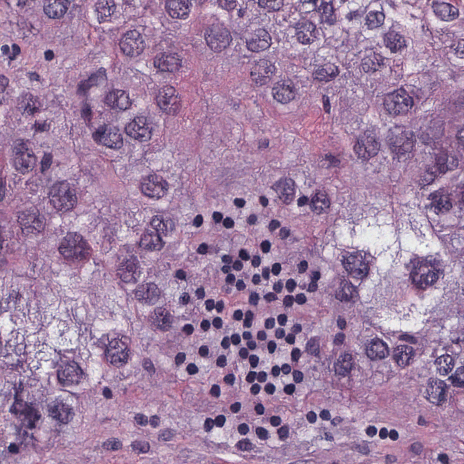}
  \end{subfigure}
    \begin{subfigure}{0.15\linewidth}
   \includegraphics[width=1\linewidth]{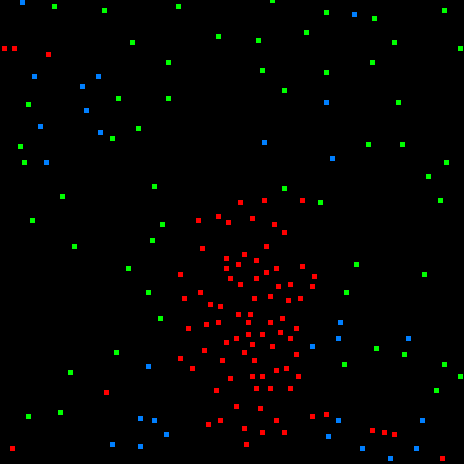}
  \end{subfigure}
    \begin{subfigure}{0.15\linewidth}
   \includegraphics[width=1\linewidth]{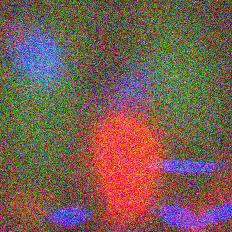}
  \end{subfigure}

% \hspace{.002in}
  \begin{subfigure}{0.15\linewidth}
   \includegraphics[width=1\linewidth]{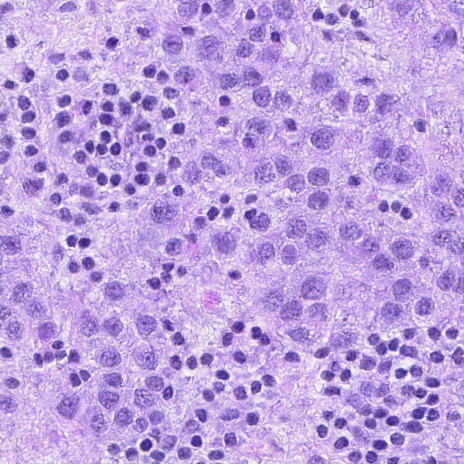}
  \end{subfigure}
  \begin{subfigure}{0.15\linewidth}
   \includegraphics[width=1\linewidth]{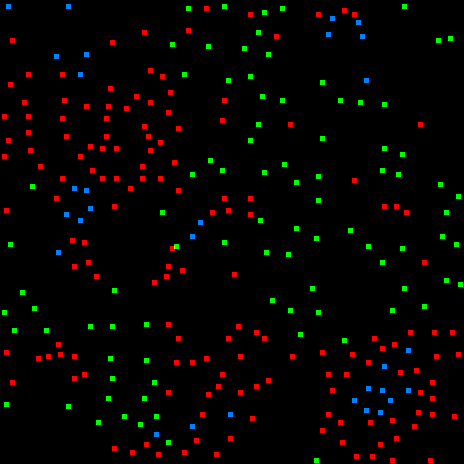}
  \end{subfigure}
  \begin{subfigure}{0.15\linewidth}
   \includegraphics[width=1\linewidth]{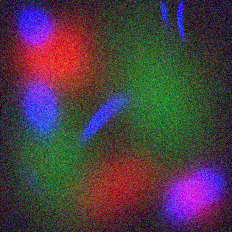}
  \end{subfigure}
      \begin{subfigure}{0.15\linewidth}
   \includegraphics[width=1\linewidth]{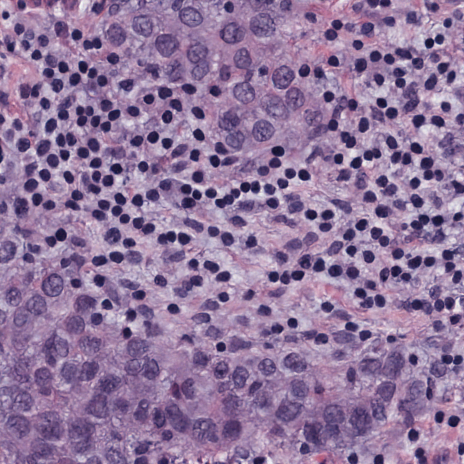}
  \end{subfigure}
    \begin{subfigure}{0.15\linewidth}
   \includegraphics[width=1\linewidth]{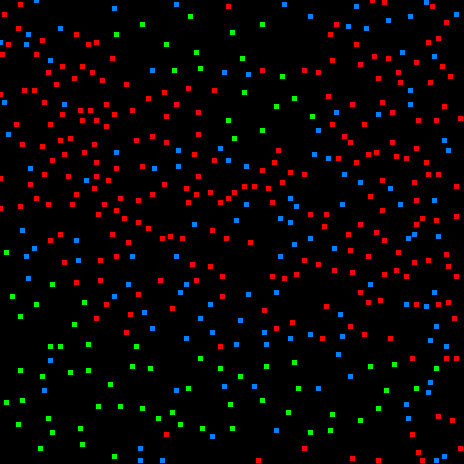}
  \end{subfigure}
    \begin{subfigure}{0.15\linewidth}
   \includegraphics[width=1\linewidth]{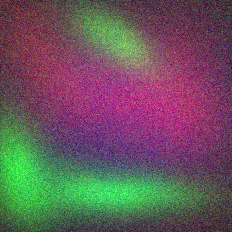}
  \end{subfigure}

\vspace{-.1in}
\caption{ The three image pairs in the left top, right top, left medium, right medium, left bottom, and right bottom are from the training set, the generation of counting categories 0, 1, 2, 3, and 4, respectively. The images in 1st and 4th columns are pathology images. The cell layouts are in 2nd and 5th columns. The generated spatial density maps are in 3rd and 6th columns.  }
% \sa{Please add legend for the colors corresponding cell types or indicate this in the caption.}} 
% \vspace{-.1in}
\label{fig:qualitative}
\vspace{-.2in}
\end{figure*}

\myparagraph{Acknowledgements}: This research was partially supported by the National Science Foundation (NSF) grant CCF-2144901, the National Institute of General Medical Sciences (NIGMS) grant R01GM148970, and the Stony Brook Trustees Faculty Award.

\myparagraph{Disclosure of Interests}: The authors have no competing interests to declare that are relevant to the content of this article.

\bibliographystyle{splncs04}
\bibliography{Paper-2613}

\begin{thebibliography}{10}
\providecommand{\url}[1]{\texttt{#1}}
\providecommand{\urlprefix}{URL }
\providecommand{\doi}[1]{https://doi.org/#1}

\bibitem{abousamra2021multi}
Abousamra, S., Belinsky, D., Van~Arnam, J., Allard, F., Yee, E., Gupta, R., Kurc, T., Samaras, D., Saltz, J., Chen, C.: Multi-class cell detection using spatial context representation. In: ICCV (2021)

\bibitem{abousamra2023topology}
Abousamra, S., Gupta, R., Kurc, T., Samaras, D., Saltz, J., Chen, C.: Topology-guided multi-class cell context generation for digital pathology. In: CVPR (2023)

\bibitem{arroyo2021variational}
Arroyo, D.M., Postels, J., Tombari, F.: Variational transformer networks for layout generation. In: CVPR (2021)

\bibitem{baranchuk2021label}
Baranchuk, D., Voynov, A., Rubachev, I., Khrulkov, V., Babenko, A.: Label-efficient semantic segmentation with diffusion models. In: ICLR (2021)

\bibitem{brock2018large}
Brock, A., Donahue, J., Simonyan, K.: Large scale gan training for high fidelity natural image synthesis. In: ICLR (2018)

\bibitem{casanova2021instance}
Casanova, A., Careil, M., Verbeek, J., Drozdzal, M., Romero~Soriano, A.: Instance-conditioned gan. In: NeurIPS (2021)

\bibitem{chai2023layoutdm}
Chai, S., Zhuang, L., Yan, F.: Layoutdm: Transformer-based diffusion model for layout generation. In: CVPR (2023)

\bibitem{choi2022perception}
Choi, J., Lee, J., Shin, C., Kim, S., Kim, H., Yoon, S.: Perception prioritized training of diffusion models. 2022 ieee. In: CVPR (2022)

\bibitem{dhariwal2021diffusion}
Dhariwal, P., Nichol, A.: Diffusion models beat gans on image synthesis. In: NeurIPS (2021)

\bibitem{gong2021style}
Gong, X., Chen, S., Zhang, B., Doermann, D.: Style consistent image generation for nuclei instance segmentation. In: WACV (2021)

\bibitem{graikos2022diffusion}
Graikos, A., Malkin, N., Jojic, N., Samaras, D.: Diffusion models as plug-and-play priors. In: NeurIPS (2022)

\bibitem{harvey2022flexible}
Harvey, W., Naderiparizi, S., Masrani, V., Weilbach, C., Wood, F.: Flexible diffusion modeling of long videos. In: NeurIPS (2022)

\bibitem{ho2020denoising}
Ho, J., Jain, A., Abbeel, P.: Denoising diffusion probabilistic models. In: NeurIPS (2020)

\bibitem{hofener2018deep}
H{\"o}fener, H., Homeyer, A., Weiss, N., Molin, J., Lundstr{\"o}m, C.F., Hahn, H.K.: Deep learning nuclei detection: A simple approach can deliver state-of-the-art results. Computerized Medical Imaging and Graphics  (2018)

\bibitem{hou2019robust}
Hou, L., Agarwal, A., Samaras, D., Kurc, T.M., Gupta, R.R., Saltz, J.H.: Robust histopathology image analysis: To label or to synthesize? In: CVPR (2019)

\bibitem{huang2023affine}
Huang, J., Li, H., Wan, X., Li, G.: Affine-consistent transformer for multi-class cell nuclei detection. In: ICCV (2023)

\bibitem{hung2020keras}
Hung, J., Goodman, A., Ravel, D., Lopes, S.C., Rangel, G.W., Nery, O.A., Malleret, B., Nosten, F., Lacerda, M.V., Ferreira, M.U., et~al.: Keras r-cnn: library for cell detection in biological images using deep neural networks. BMC bioinformatics  (2020)

\bibitem{inoue2023layoutdm}
Inoue, N., Kikuchi, K., Simo-Serra, E., Otani, M., Yamaguchi, K.: Layoutdm: Discrete diffusion model for controllable layout generation. In: CVPR (2023)

\bibitem{jiang2022coarse}
Jiang, Z., Sun, S., Zhu, J., Lou, J.G., Zhang, D.: Coarse-to-fine generative modeling for graphic layouts. In: AAAI (2022)

\bibitem{jyothi2019layoutvae}
Jyothi, A.A., Durand, T., He, J., Sigal, L., Mori, G.: Layoutvae: Stochastic scene layout generation from a label set. In: ICCV (2019)

\bibitem{karras2019style}
Karras, T., Laine, S., Aila, T.: A style-based generator architecture for generative adversarial networks. In: CVPR (2019)

\bibitem{kasa2022improved}
Kasa, S.R., Rajan, V.: Improved inference of gaussian mixture copula model for clustering and reproducibility analysis using automatic differentiation. Econometrics and Statistics  (2022)

\bibitem{kikuchi2021constrained}
Kikuchi, K., Simo-Serra, E., Otani, M., Yamaguchi, K.: Constrained graphic layout generation via latent optimization. In: ACM MM (2021)

\bibitem{li2020layoutgan}
Li, J., Yang, J., Hertzmann, A., Zhang, J., Xu, T.: Layoutgan: Synthesizing graphic layouts with vector-wireframe adversarial networks. PAMI  (2020)

\bibitem{luo2021diffusion}
Luo, S., Hu, W.: Diffusion probabilistic models for 3d point cloud generation. In: CVPR (2021)

\bibitem{miyato2018cgans}
Miyato, T., Koyama, M.: cgans with projection discriminator. In: ICLR (2018)

\bibitem{nichol2021improved}
Nichol, A.Q., Dhariwal, P.: Improved denoising diffusion probabilistic models. In: ICML (2021)

\bibitem{odena2017conditional}
Odena, A., Olah, C., Shlens, J.: Conditional image synthesis with auxiliary classifier gans. In: ICML (2017)

\bibitem{reynolds2009gaussian}
Reynolds, D.A., et~al.: Gaussian mixture models. Encyclopedia of biometrics  (2009)

\bibitem{ronneberger2015u}
Ronneberger, O., Fischer, P., Brox, T.: U-net: Convolutional networks for biomedical image segmentation. In: MICCAI (2015)

\bibitem{saharia2022image}
Saharia, C., Ho, J., Chan, W., Salimans, T., Fleet, D.J., Norouzi, M.: Image super-resolution via iterative refinement. PAMI  (2022)

\bibitem{scott:1992}
Scott, D.W.: Multivariate Density Estimation. Theory, Practice, and Visualization. Wiley (1992)

\bibitem{sohl2015deep}
Sohl-Dickstein, J., Weiss, E., Maheswaranathan, N., Ganguli, S.: Deep unsupervised learning using nonequilibrium thermodynamics. In: ICML (2015)

\bibitem{sugimoto2022multi}
Sugimoto, T., Ito, H., Teramoto, Y., Yoshizawa, A., Bise, R.: Multi-class cell detection using modified self-attention. In: CVPR (2022)

\bibitem{tewari2023estimation}
Tewari, A.: On the estimation of gaussian mixture copula models. In: ICML (2023)

\bibitem{tsai2020adhesion}
Tsai, T.Y.C., Sikora, M., Xia, P., Colak-Champollion, T., Knaut, H., Heisenberg, C.P., Megason, S.G.: An adhesion code ensures robust pattern formation during tissue morphogenesis. Science  (2020)

\bibitem{yousefi2019transfer}
Yousefi, S., Nie, Y.: Transfer learning from nucleus detection to classification in histopathology images. In: ISBI (2019)

\bibitem{zhang2019self}
Zhang, H., Goodfellow, I., Metaxas, D., Odena, A.: Self-attention generative adversarial networks. In: ICML (2019)

\bibitem{zhu2022fine}
Zhu, H., Yuan, J., Yang, Z., Zhong, X., Wang, Z.: Fine-grained fragment diffusion for cross domain crowd counting. In: ACMMM (2022)

\end{thebibliography}
\end{document}